# Transformer-based toxin-protein interaction analysis prioritizes airborne particulate matter components with potential adverse health effects

*Yan Zhu*[1,2,3], *Shihao Wang*[4], *Yong Han*[4], *Yao Lu*[4], *Shulan Qiu*[1,2], *Ling N. Jin*[2,4], *Xiangdong Li*[4], *Weixiong Zhang*[1,2,5,†]

1. Hong Kong Jockey Club STEM Laboratory of Genomics and AI in Healthcare, The Hong Kong Polytechnic University, Hong Kong, China
2. Department of Health Technology and Informatics, The Hong Kong Polytechnic University, Hong Kong, China
3. Faculty of Computing, Harbin Institute of Technology, Harbin 150001, China
4. Department of Civil and Environmental Engineering, The Hong Kong Polytechnic University, Hong Kong, China
5. Department of Computing, The Hong Kong Polytechnic University, Hong Kong, China

†. Correspond to: weixiong.zhang@polyu.edu.hk

**Abstract**

Air pollution, particularly airborne particulate matter (PM), poses a significant threat to public health globally. It is crucial to comprehend the association between PM-associated toxic components and their cellular targets in humans to understand the mechanisms by which air pollution impacts health and to establish causal relationships between air pollution and public health consequences. Current methods for modeling and analyzing these interactions are rudimentary, with experimental approaches offering limited throughput and comprehensiveness. Leveraging cutting-edge deep learning technologies, we developed *tipFormer* (*t*oxin-protein *i*nteraction *p*rediction based on trans*former*), a novel machine-learning approach for identifying toxic components capable of penetrating human cells and instigating pathogenic biological activities and signaling cascades. It incorporates dual pre-trained language models to derive encodings for protein sequences and chemicals. It employs a convolutional encoder to assimilate the sequential attributes of proteins and chemicals. It then introduces a novel learning module with a cross-attention mechanism to decode and elucidate the multifaceted interactions pivotal for the hotspots binding proteins and chemicals. Through thorough experimentation, tipFormer was shown to be proficient in capturing interactions between proteins and toxic components. This approach offers significant value to the air quality and toxicology research communities by enabling high-throughput, high-content identification and prioritization of hazards. It facilitates better-targeted wet lab mechanistic studies and field measurements, ultimately advancing our understanding of air pollution's impact on human health.

## 1. Introduction

Air pollution has emerged as a critical global health concern, primarily driven by rapid economic, industrial and population growth and further exacerbated by climate change and other non-anthropogenic factors [1]. The World Health Organization estimates that approximately 7 million premature deaths occur every year due to air pollution exposure. The consequences of air pollution extend far beyond individual health implications and exacerbate the strain on societal and healthcare systems in numerous ways [2]. The health risks associated with airborne particulate matter (PM) are particularly concerning for public health [3]. These particles emanate from a wide spectrum of sources, encompassing both anthropogenic origins—such as biomass burning and vehicular emissions—and natural sources, including windblown dust, volcanic eruptions, and vegetation [4]. Exposure to PM has been strongly linked to a wide range of adverse health effects beyond just respiratory issues [5], including cardiovascular diseases [6], neurodegenerative disorders [7], and pregnancy complications [8]. The ability of fine and ultrafine PM to penetrate deep into the body, cross biological barriers, and cause systemic inflammation and oxidative stress is thought to contribute to these wide-ranging health effects [9]. These reported health consequences highlight the importance and urgency of air pollutants control. However, current control strategies often tend to focus on controlling the mass concentration of PM, which has significant uncertainty in improving health risks. The primary reason is the non-equivalence between PM mass and their toxicity effects. This necessitates the identification of key toxic components in airborne PM on the one hand and the understanding of the mechanisms of these key toxic components, on the other hand, in order to understand their association with diseases. A comprehensive investigation into the molecular mechanisms underlying the toxicity induced by air pollutants is imperative to address this multifaceted issue.

Over the past decades, significant efforts have been invested in understanding the intricate mechanisms of toxicity of PM through *in vitro*, *in vivo*, and epidemiological studies. These studies focus on assessing the biological and health impacts of PM. Moreover, numerous studies have explored the toxicological mechanisms of PM through *in vitro* and *in vivo* experiments. *In vitro* studies using cell culture models are employed to assess the cytotoxicity of PM. These studies utilize various assays to measure the effects of PM exposure on cell viability, oxidative stress, and inflammatory responses [10]. The commonly used cell types in these studies are lung epithelial cells, macrophages, and organoids [11]. *In vivo* studies involve animal exposure experiments through inhalation or instillation, allowing identification and analysis of biomarkers of exposure and effect in blood, urine, and tissue samples [12].

While experimental methods offer invaluable insights, they are often time-consuming and costly, particularly when conducting exhaustive studies required for a full spectrum of chemicals and mixtures across the full spectrum of biological endpoints and pathways. In contrast, *in silico* approaches offer a way for accelerated prioritization of what to test experimentally. Computational methods have proven effective in accelerating drug discovery and enabling precision medicine [13, 14]. They can also offer a promising way of identifying potential toxin-target interactions (TTIs) and assessing toxicity. Such computational approaches are supported by the increasing availability of toxic component data and omics data, including genomic, proteomic, and metabolomic data. Furthermore, machine learning (ML) methods can be integrated with experimental techniques [15]. This integration enables high-throughput outcomes, cost reduction, time efficiency enhancement, and minimization of unintentional experimental errors. Overall, the integration of massive data, machine learning, and experimental techniques provides a robust framework for enhancing our understanding of toxin-target interactions and toxic mechanisms.

In sharp contrast to the large body of literature on predicting protein-protein interactions [16-20], to the best of our knowledge, only one study has adequately exploited ML methods for predicting TTIs [21]. In this study, traditional ML algorithms – namely, support vector machines (SVM) and random forests (RF) – were employed to develop classification models to predict these interactions. The study extracted a set of 1,547 molecular features for toxins and 400 features for target proteins, which served as inputs to the SVM and RF classifiers. The results indicated positive potential. The proposed approach showed high accuracy in predicting TTIs. However, it has notable limitations. The study was restricted to a specific collection of toxin-target pairs, which may not be generalizable to diverse data types, and it cannot explain or interpret the learned features and prediction outcomes, limiting its applicability and reliability.

Deep learning (DL), a rapidly expanding subfield of machine learning, has delivered exceptional performance across a wide range of applications in various domains, including computer vision [22], natural language processing [23], and drug discovery [24]. Unlike traditional ML methods, DL can automatically learn intricate patterns and representations from raw data, which helps in making more accurate predictions. This is particularly useful for predicting toxin-target interactions (TTIs) because DL can handle diverse and complex data types more effectively. One of the key advancements in DL is the use of transformers, which have revolutionized how we understand and represent data [25]. Transformers can capture long-range relationships within data, making them valuable for tasks that require a deep understanding of complex interactions. For example, pre-trained models like BERT [26], ELECTRA [27], and T5 [28], built on transformer architecture, have shown remarkable success in various applications. These models can potentially provide more reliable and interpretable predictions for TTIs by learning from a broader range of data and offering insights into the features that drive their predictions.

Given the success of transformer-based DL in various application domains, we postulate that DL, particularly transformers, can be readily extended to the prediction of TTIs as an alternative to experimental tests of the toxicity of air pollutants. We developed *tipFormer* (*t*oxin-protein *i*nteractions *p*rediction based on trans*former*), a novel transformer-based deep learning framework for predicting interactions between atmospheric pollutants and protein targets. TipFormer was designed to predict such interactions using the sequences of protein amino acids and strings of chemical compounds represented in Simplified Molecular Input Line Entry Specification (SMILES). Specifically, we first constructed comprehensive feature profiles of chemical compounds and protein sequences based on pre-trained chemical language models (CLMs) and pre-trained protein language models (PLMs), respectively. We then adopted a pair of convolutional neural networks (CNNs) to learn abstract, low-dimensional joint representations of compounds and proteins. Finally, we integrated compound and protein interrelationships through a cross-attention mechanism to discern their intricate interactions. An explanation mechanism was introduced to identify the most important compound and protein features that facilitate the interactions between the two. The potential cellular and molecular influence of the identified toxins was examined by the biological processes and signaling pathways that the interacting proteins may perturb. As an application, we applied tipFormer to assess the potential health impact of an ensemble of typical organic components present in source-specific and ambient PM.

## 2. Results

We developed tipFormer, an *in silico* approach, to identify components in PM pollutants that may disrupt cellular processes resulting in health problems. Tipformer is based on transformer, the latest deep learning technique for extracting and analyzing key features in large data. Complementary to *in vitro* experimental analysis, the new method focuses on individual chemical compounds and is scalable to large datasets.

### 2.1. TipFormer and its characteristics and performance

#### 2.1.1. The architecture and approach

TipFormer is a transformer-based deep-learning approach to learning and predicting the interaction between aerosol toxins and protein targets. The tipFormer method was built upon the powerful encoder-decoder framework of the Transformer, which consists of an encoder and a decoder utilizing the self-attention mechanism with feed-forward networks [25]. We considered the sequences of proteins and toxin molecules as linguistic strings, leveraging the superior capabilities of transformer-based architecture to process these chemical and biological sequences. The resulting model was able to learn complex interactions between toxins and proteins accurately.

TipFormer is composed of six modules (see Methods and Figure 1): (1) Data collection – consisting of the T3DB database as the foundational basis (Figure 1A); (2) Preprocessing – selecting air pollutant organic molecules with high quality from T3DB (Figure 1B); (3) The encoding module – employing pre-trained language models enhanced with convolutional layers to extract features of protein sequences and toxin SMILES strings (Figure 1C); (4) The interaction learning module – introducing a cross-domain attention mechanism to assimilate interactions between toxins and proteins (Figure 1D); (5) The pairwise interaction prediction module – applying acquired features to predict interactions between toxins and proteins (Figure

1E); and (6) Model evaluation and interpretation – ensuring the model's interpretability and generalizability (Figure 1F).

### 2.1.2. Characteristics of tipFormer

#### 2.1.2.1. TipFormer delivers preferable performance

We compared tipFormer with four benchmark methods, including three conventional machine learning methods and a deep learning-based (DL-based) method. Due to the possible influence of negative samples and random parameter initialization, we assessed the performance of each method five times. We reported the average results of accuracy (Acc), sensitivity (Sn), specificity (Sp), precision (Pre), F1 score, and Matthew's correlation coefficient (MCC) (see Methods, Figure 2A, Table 1, and Supplementary Table S1). We also considered the overall performance of ROC curves and AUC values under the predictions of tipFormer (Figure 2B).

Among the five methods compared, the two DL-based methods, tipFormer and DeepCNN, exhibit the best and second-best accuracies of 0.895 and 0.887, respectively, showing their ability to identify and categorize interacting toxin-target pairs accurately. Importantly, tipFormer outperforms the other methods not only in accuracy but also in the two most important overall performance measures, F1 and MCC (Figure 2A, Table 1). It also performs admirably well in the other performance measures. The preferable performance of tipFormer is likely due to its ability to autonomously learn complex feature representations of data, thereby more accurately capturing the intrinsic patterns and regularities within the data. The new cross-attention mechanism introduced in tipFormer (see Methods) enables it to extract complex interactions between proteins and toxins adeptly.

Although the KNN method achieves the best specificity and precision, its accuracy and overall performance measures of F1 score and MCC are inferior to tipFormer and DeepCNN (Figure 2A, Table 1). This may be in part due to the limitations of KNN in dealing with complex data distributions and feature relationships. Note that all the methods compared have small errors, suggesting that they are relatively stable for TTI prediction. This rigorous comparison enables us to assess the performance of our method against a diverse set of established prediction techniques, providing a solid foundation for evaluating the effectiveness and reliability of our approach.

We examined and visualized how tipFormer mapped the input data to a lower-dimensional space using t-SNE (t-distributed Stochastic Neighbor Embedding) [29] to the input and output data of tipFormer (Figure 2). The data features are scattered in the input space (Figure 2C) but converge to form distinct clusters in the low-dimensional representation space (Figure 2D), illustrating tipFormer's classification prowess.

#### 2.1.2.2. TipFormer is generalizable

An essential feature of a prediction method is its generalizability to different, albeit similar, application scenarios. For the problem of predicting TTIs, it is important to identify all putative interactions between a given set of toxins and proteins where some of the toxins, some of the proteins, or both have never been exposed to the prediction model. Note that these three scenarios are different from the case that we considered earlier in building classification models. We considered the two scenarios here and left the third one in the next section.

We regenerated training, validation and test data using two new-split settings of the large benchmark dataset we collected [30, 31]. The new-split settings mimic real-world situations where models handle toxins or targets that are absent from the training set, thereby comprehensively assessing their generalizability. We considered the following two scenarios:
    (1) New-toxin setting: no toxin was used in the training and test sets;
    (2) New-target setting: no target was used in the training and test sets.

We evaluated the performance of tipFormer in comparison with the five benchmark methods considered previously (Figures 2E, 2F and Tables S2, S3). As expected, each model exhibited varying degrees of deteriorating performance in the new scenarios, yet tipFormer was the most robust. TipFormer achieved the highest accuracy scores of 0.639 and 0.653 under the new-toxin setting (Figure 2E) and new-target setting (Figure 2F), respectively. While the SVM model showed excellent sensitivity (Sn value of 1), indicating its high efficacy in identifying true interactions, it had the lowest specificity (Sp value of 0). This indicates a significant shortfall in the model's capacity to differentiate non-interacting samples, which could result in a high rate of false positives. The RF model also displayed high sensitivity in the new-toxin setting, but similarly, its specificity was the lowest (Figure 2E). Taking all metrics into account, tipFormer performed superiorly to the existing methods, making it the overall preferred method for TTI prediction.

### 2.1.2.3. TipFormer is generalizable to new data

We compiled an independent test dataset of 20 potential toxic compounds from a broad range of sources of airborne particulate matter (ambient air, biomass burning, traffic sources, and marine aerosols) based on non-targeted screening to test our model. It is important to note that these 20 potential toxic compounds were not included in the previous database. We selected 101 proteins from the PubChem database [32] that interact with these toxins, for a total of 138 interactions between the toxins and proteins (Supplementary Table S4). Note that neither these toxins nor proteins are in the benchmark dataset. The comparison results (Table 2) showed that the new model achieved a precision rate of 0.790 in predicting the 138 interactions between 20 types of toxins and 101 proteins. While the SVM algorithm predicted all interactions as positive, it had a propensity for overfitting to the positive samples, as shown in the previous section. Furthermore, an intriguing finding is that the DeepCNN model, which also employs deep learning, performed second next to tipFormer. This result showed the superior efficacy of deep learning approaches over traditional machine learning methods for this task. These results validated the robustness and generalizability of tipFormer in predicting the interaction between new toxins and new proteins. We list the predicted values for these 101 interactions in Supplementary Table S5.

### 2.2. TipFormer identifies components of ambient and source-specific PM with potential adverse health impacts

It has been proven that the key toxic components in inhalable particulate matter are one of the important factors causing health risks in the population. Therefore, an urgent issue to be addressed is how to screen for these key toxic components and understand how the toxic components from different sources of aerosols interact with various proteins, thereby affecting human health.

We compiled a dataset of 94 compounds generated from biomass burning, vehicular exhaust, and ambient air. To test if these toxins can interact with human proteins and consequently cause health issues, we extracted 20,434 high-quality, manually curated human proteins from the UniProt database [33]. TipFormer plus statistical analysis predicted that the 94 toxins interact with 955 proteins; the results are on GitHub (https://github.com/YanZhu06/tipFormer). The significant enrichment in glutathione transferase activity indicates that the antioxidant defense mechanism is compromised. This is consistent with previous studies [34-36] that attributed the adverse health impacts from airborne PM to the oxidative stress. This is also consistent with our molecular functional analysis of the gene ontology of these interacting proteins, further validating the accuracy of the model predictions from a culture perspective. Our analysis is presented below.

To investigate the implications of these interactions, we conducted a gene-disease association analysis using the GeDiPNet database [37], a comprehensive source of information on human genes, single nucleotide polymorphisms (SNPs), and diseases. The analysis revealed diseases that may be induced by some of the interactions between the toxins and proteins with a statistical significance of $p$-value no greater than 0.05 (Table 4). Notably, neurological diseases are the most prevalent, indicating a strong correlation with the predicted toxin-interacting proteins and their corresponding genes. Neurodegenerative diseases of the central nervous system ($p$-value = 0.010), neurodegenerative disorders ($p$-value = 0.010), cryptogenic tonic-clonic epilepsy ($p$-value = 0.010), tonic-clonic epilepsy ($p$-value = 0.017), and schizophreniform disorders ($p$-value = 0.027) all showed highly significant associations. These findings suggest that gaseous toxins from biomass combustion may contribute to the development or exacerbation of neurological diseases, potentially through interactions with specific proteins and dysregulation of associated genes and pathways. In addition, cardiovascular diseases such as pulmonary hypertension ($p$-value = 0.011) and secundum atrial septal defect ($p$-value = 0.032) also exhibited statistically pronounced $p$-values, indicating a potential link between these gaseous toxins and cardiovascular disease development. Other disease categories, including ocular diseases (cortical cataract, $p$-value = 0.021), hepatobiliary diseases (intrahepatic cholestasis, $p$-value =

0.038), and congenital malformations (isolated split hand-split foot malformation, *p*-value = 0.041) were also statistically discernible.

These findings are concordant with previous human and animal studies showing that air pollution, particularly that of PM, is closely associated with neurodegenerative diseases such as Alzheimer's and Parkinson's, as well as cardiovascular diseases like ischemic heart disease, stroke, hypertension, and heart failure [38-40]. Our results further validated the superior performance of tipFormer in predicting interactions between air pollutants and proteins, offering tipFormer as an enabling tool for understanding and addressing the impact of PM pollution on public health.

We attempted to gain a deep insight into the mechanisms by which PM may contribute to neurodegenerative diseases. To this end, we performed a GO analysis on the 955 proteins (or genes) that interact with the 94 OA toxins to identify the biological functions and signaling pathways that these toxins target (Figure S1).

The GO analysis revealed several significant biological processes that may be important for neurodegenerative diseases (Figure S1A). Specifically, Vesicle Fusion with Golgi Apparatus (GO:0048280) and Vesicle Fusion (GO:0006906) are statistically significant with extremely low *p*-values, indicating a potential disruption in intracellular transport mechanisms, which are crucial for neuronal function and survival [41, 42]. Additionally, Positive Regulation of Transcription Initiation by both DNA-templated transcription initiation (GO:2000144) and RNA Polymerase II (GO:0060261) also demonstrate statistical relevance, suggesting that toxins may interfere with gene expression regulation, impacting neuronal health [43]. Cilium Movement Involved in Cell Motility (GO:0060294) and Intestinal Epithelial Cell Differentiation (GO:0060575) have low *p*-values, hinting at a broader impact on cellular movement and differentiation relevant to neurodegenerative processes [44]. Activation of Cysteine-Type Endopeptidase Activity Involved in the Apoptotic Process by Cytochrome C (GO:0008635) shows marked enrichment, indicating a potential induction of apoptosis by the toxins, a process often implicated in neurodegeneration [45, 46].

The cellular component analysis pinpointed specific cellular components and structures that may be affected by the OA toxins (Figure S1B). Core Mediator Complex (GO:0070847) and U6 snRNP (GO:0005688) showed significant associations, suggesting that the toxins may impact RNA processing and splicing, which are essential for maintaining neuronal function. Coated Vesicle (GO:0030135) and COPII-coated ER to Golgi Transport Vesicle (GO:0030134) were also implicated, indicating potential disruptions in vesicle-mediated transport, a critical process for neuronal communication. Spliceosomal Complexes, including the Precatalytic Spliceosome (GO:0071011) and U2-type Spliceosomal Complex (GO:0005684), were highlighted, further emphasizing the potential impact on RNA splicing and gene expression regulation.

The molecular function analysis identified key activities that the toxins may influence (Figure S1C). Sequence-Specific Double-Stranded DNA Binding (GO:1990837) and Double-Stranded DNA Binding (GO:0003690) were significantly enriched, suggesting that toxins may directly interact with DNA to affect its integrity. Glutathione Transferase Activity (GO:0004364) and Glutathione Binding (GO:0043295) were also significant, indicating a potential disruption in the antioxidant defense mechanisms, which are vital for combating oxidative stress associated with neurodegeneration. Transcription Regulatory Region Nucleic Acid Binding (GO:0001067) was implicated, further supporting the notion that toxins may interfere with gene expression regulation.

Among these interacting proteins, 85 (or 8.9% of the 955) are cell membrane proteins, suggesting that they may bind with toxin components to accommodate their entry into the cell. To explore the role of the interactions between the 94 toxins and 85 cell membrane proteins in promoting disease development and progression, we also analyzed these 85 proteins using the GeDiPNet database. Table 3 presents the outcomes of this analysis, highlighting associations between cell membrane proteins and diseases, with a *p*-value threshold set at no more than 0.05, denoting statistical significance. The data suggests a pivotal role for cell membrane proteins in the development of various diseases, particularly those affecting the urinary system, skin, and nervous system. Notably, the table lists diseases potentially associated with environmental factors, such as toxic shock syndrome [47] and arsenic encephalopathy [48]. This finding strongly suggests that these membrane proteins may play a crucial role in diseases induced by environmental toxins.

We performed GO enrichment analyses of the identified cell membrane proteins to elucidate their potential roles in mediating environmental toxin effects (Figure S2A-C). The analysis revealed significant enrichment of water and small molecule transport-related functions across all three GO categories. Notably, water channel activity (GO:0015250) and fluid transport (GO:0042044) were among the most significantly enriched terms (Figure S2A, S2C). This enrichment suggests a crucial role for these proteins in cellular osmotic regulation and homeostasis, consistent with previous findings [49]. Membrane-associated functions were prominently represented, as evidenced by the enrichment of basolateral plasma membrane (GO:0016323) in CC (Figure S2B) and various channel activities in MF (Figure S2C). These findings align with the expected functions of cell membrane proteins and underscore their importance in cell-environment interactions [50]. The analysis also highlighted enrichment in cellular signaling and endocytosis processes, including G protein-coupled receptor internalization (GO:0002031) and components related to clathrin-coated vesicles (Figure S2A, S2B). This implies the potential involvement of these proteins in signal transduction and substance uptake mechanisms, which could be particularly relevant to the cellular processing of environmental toxins. The diverse array of enriched functions spans from specific molecular transport to broader cellular processes like cell adhesion (GO:0007155) and developmental processes such as odontogenesis (GO:0042476), indicating the multifaceted roles these proteins may play in various physiological contexts (Figure S2A). The strong emphasis on water and small molecule transport, membrane-associated functions, and signaling pathways provides valuable clues about potential mechanisms by which environmental toxins might interact with and affect cellular systems.

In short, the toxins we analyzed interact with and affect the functions of many proteins that are involved in key cellular processes, such as RNA splicing, membrane transport, DNA binding, redox metabolism, and transcriptional regulation. Such interactions may alter the functions of these proteins to exert their biological effects by affecting these processes and molecular functions, resulting in adverse consequences like neurological disorders, pulmonary diseases, and visual impairments. These findings provide mechanistic into the function and action of the toxins from organic biomass combustion on neurodegenerative diseases.

### 2.3. TipFormer discovers hotspot protein residues critical for interaction

TipFormer computes the outer product of the feature matrices as shown in Figure 1D (the Interaction Learning Module), thereby constructing a unique attention map. Visualizing this attention map helps clearly identify which parts of the protein should receive greater attention for interaction with toxins.

We refer to those residues that are most closely engaged in interaction with toxins as "hotspot residues." Take the compound Bisphenol A (BPA) as an example. It is a bisphenol compound structurally characterized as 4,4'-methanediyldiphenol (Figure 3A). BPA plays multiple roles in the ecosystem, functioning as an exogenous estrogen, an environmental contaminant, a xenobiotic, and an endocrine disruptor [51, 52]. Extensively utilized in the production of polycarbonate plastics, epoxy resins, and non-polymers, BPA can interact with nuclear receptors, thereby disrupting the normal function of the endocrine system. As an endocrine-disrupting chemical (EDC), BPA has the potential to induce cancer and obesity and reduce fertility [53, 54]. BPA's interaction with the Estrogen-related receptor gamma (ERRγ) is significant, as evidenced by the protein residues involved in this interaction (Figure 3B). These residues, numbered 268, 272, 275, 306, 309, 316, 322, 323, 324, 326, 332, 333, 334, 337, 343, 345, 346, 347, 349, 350, 390, 394, 435, 439, 442, 444, 445, 446, engage with the ligand (Figure 3C) [55, 56]. According to the attention matrix generated by the tipFormer model, the top 28 residues are 358, 390, 303, 298, 282, 309, 413, 388, 412, 439, 306, 338, 372, 442, 332, 343, 328, 334, 404, 415, 437, 308, 366, 260, 317, 367, 417, 312. Among these, eight residues coincide exactly with the interacting residues listed above, and many other residues, such as 308, 317, 338, 328, and 437, are in close proximity to the hotspot residues. This result provides valuable insights into toxin-protein interactions, suggesting the potential of our model in identifying hotspot residues that interact with ligands.

### 3. Materials and Methods

### 3.1. The aerosol pollutant dataset

We constructed a dataset of aerosol pollutant toxins using the data in the Toxin and Toxin-Target Database (T3DB, http://www.t3db.org). This comprehensive repository provides data on various types of toxins, their targets, and the intricate interactions between the two [57]. Within the database is detailed mechanistic information on protein-toxin interactions. For instance, Dichlorvos, aka 2,2-dichlorovinyl dimethyl phosphate

(DDVP), is a highly volatile organophosphate insecticide widely employed to control household pests such as spider mites and caterpillars in fruits and vegetables [58]. However, Dichlorvos disrupts the human endocrine system by binding to and inhibiting the estrogen receptor (Target UniProt ID: P03372), posing potential health risks [59]. The T3DB currently contains 3,678 toxins, which are described by 41,602 synonyms. These toxins include various organic and inorganic substances that are airborne pollutants, pesticides, drugs, and cigarette and food toxins. They are linked to 2,073 toxin targets, giving rise to a total of 42,374 toxin-target associations (Figure 1A).

We were interested in organic toxins in air pollutants. We thus screened the T3DB for data related to air pollutants and filtered the entries based on specific criteria. The selected compounds were required to have valid SMILES representations in PubChem [32], and chosen proteins needed to be mapped to UniProt IDs with sequence data [33]. The resulting dataset (available at https://github.com/YanZhu06/tipFormer) comprises 446 toxins, 719 associated proteins, and 8,206 interactions for further analysis.

The analysis of the potential health impact of this dataset can be found in Supplementary Text S1 and Figure S3. The analysis encompasses the utilization of the Gene Ontology (GO) and Kyoto Encyclopedia of Genes and Genomes (KEGG) databases for enrichment analysis, aiming to uncover potential interactions between these genes and air pollutant toxins. Additionally, the distribution of protein sequences and toxin SMILES strings are also included in the analysis. We randomly split this baseline dataset into three subsets: a training set for model building, a validation set for hyperparameter tuning, an optimal model saving set, and an independent test set for model evaluation.

### 3.2. Sequence embeddings and representation

We employed two transformer-based language models to derive sequence feature vectors for toxin molecules and protein sequences. For the toxin molecules, we utilized ChemBERTa [60], a large chemical language model for embedding and representing molecule structural properties and relationships. ChemBERTa is trained in a self-supervised manner on SMILES strings corresponding to a large collection of chemical molecules in the public database PubChem [32]. This model is specifically designed with an efficient attention mechanism with positional embeddings [61], aiming to learn a compressed representation of chemical molecules. In tipFormer, we exploited the molecular embeddings in ChemBERTa as a foundational layer to capture the intricate details and nuances of toxin molecules (Figure 1C). The embeddings were then used to inform subsequent layers of tipFormer.

For protein amino acid sequences, we made use of ProtBert [62], which embeds protein structural properties into a low-dimensional representation. As its name indicates, ProtBert is a model based on the bidirectional encoder representations from transformers (BERT) [26], which has been pre-trained exclusively on protein sequences using a masked language modeling (MLM) objective. This model is adept at understanding the 'language' of proteins, as it has been trained on raw protein sequences with no label, relying on an automated process to generate inputs and labels from these sequences. The embeddings produced by ProtBert capture important biophysical properties that govern protein three-dimensional structures, suggesting that the model has learned aspects of the grammar of life encoded in protein sequences [62]. We took advantage of the ProtBert embeddings in tipFormer to encapsulate the contextual information and structural and semantic relationships of protein sequences.

Given the SMILES strings of toxins $t_i = \{t_i^1, t_i^2, \ldots, t_i^n\}$, and the amino acid sequences of target proteins $p_i = \{p_i^1, p_i^2, \ldots, p_i^m\}$, the embedding representations of the toxins and the proteins can be written as

$$X_t = \text{CLM}(t_i), X_t \in \mathbb{R}^{n \times D_t} \quad (1)$$

$$X_p = \text{PLM}(p_i), X_p \in \mathbb{R}^{n \times D_p} \quad (2)$$

where CLM() and PLM() represent ChemBERTa and ProtBert, and $D_t$ and $D_p$ have 384 and 1,024 dimensions for the embeddings of the toxins and the targets, respectively.

### 3.3. The components of tipFormer

#### 3.3.1. The encoder module

Given the sequence representations of toxins and proteins from the CLM and PLM, we refrained from feeding them into the decoder directly. Instead, we further encoded these initial sequence representations to a lower-dimension space to learn more abstract features (Figure 1C). The toxin representation $X_t$ and the protein representation $X_p$ were encoded by linear layers to derive the embedding vectors $X_t^e$ and $X_p^e$, respectively:

$$X_t^e = W_t^{(1)} X_t + b_t^{(1)} \qquad (3)$$

$$X_p^e = W_p^{(1)} X_p + b_p^{(1)} \qquad (4)$$

where $W_t^{(1)}$ and $W_p^{(1)}$ are the learnable weight matrices, and $b_t^{(1)}$ and $b_p^{(1)}$ are the learnable bias vectors in the linear layers.

The resultant features were subsequently updated through typical 1-D convolution layers employing a Gated Linear Unit (GLU) activation function to obtain $H_t^{(1)}$ and $H_p^{(1)}$.

$$H_t^{(1)} = \sigma(\text{CNN}(W_t^{(2)}, b_t^{(2)}, X_t^e)) \qquad (5)$$

$$H_p^{(1)} = \sigma(\text{CNN}(W_p^{(2)}, b_p^{(2)}, X_p^e)) \qquad (6)$$

where $W_t^{(2)}$ and $W_p^{(2)}$ are the learnable weight matrices, $b_t^{(2)}$ and $b_p^{(2)}$ are the learnable bias, and $\sigma(\ )$ denotes GLU activation function.

Finally, the layer normalization function was applied to obtain the feature vectors $H_t^{(2)}$ and $H_p^{(2)}$ of the encoder:

$$H_t^{(2)} = \text{LayerNorm}(H_t^{(1)}) \qquad (7)$$

$$H_p^{(2)} = \text{LayerNorm}(H_p^{(1)}) \qquad (8)$$

### 3.3.2. The model for learning interaction

We focused on learning the interactions between toxin molecules and proteins after obtaining the aforementioned feature maps. An imminent feature of this learning module is the cross-representation learning across the toxin molecules and proteins. This was achieved by a novel cross-attention mechanism (Figure 1D). This learning module consists of multiple stacked decoder layers, each of which contains a multi-head self-attention sublayer and a position feedforward network sublayer. The self-attention sublayer is responsible for capturing the dependencies within the sequence, and the position feedforward network is used to nonlinearly transform the output of the self-attention layer to extract features further. Furthermore, the inclusion of a cross-attention sublayer facilitates the computation of cross-attention between toxins and proteins, effectively capturing their complex interdependencies and enhancing the representation of their respective features.

The attention mechanism takes three inputs: the keys (K), the values (V), and the queries (Q). The dot product of Q and K is calculated to obtain the attention weight matrix, and then this matrix is multiplied by V to compute the attention (head), written as:

$$head_i = \text{Attention}_i = \text{Softmax}\left(\frac{Q_i K^T_{\ i}}{\sqrt{d_k}}\right) V_i, i = 1,2,\dots,h \qquad (9)$$

where $d_k$ is a scaling factor depending on the layer size, $K^T_{\ i}$ represents the transposed matrix of K, and $h$ denotes the dimension of the input.

The multi-head attention is expressed as follows:

$$\text{Multihead} = \text{concat}(head_i, \dots head_i) \cdot W \qquad (10)$$

where W is the learnable weight matrices.

After the convolutional layers produce feature maps for toxins and proteins, a self-attention layer is introduced to effectively capture the dependencies within the convolutional feature map of toxins, treating the feature map as Q, K, and V, respectively. Subsequently, a cross-attention mechanism is employed to model the interactions between toxins and proteins. The bidirectional information exchange between the attention keys and values of proteins and the attention queries of toxins facilitates the association of information between the two, capturing the interactive features between toxins and protein targets. By stacking multiple such layers, the model is able to abstract and refine the information in the input sequences layer by layer, generating higher-level representations. In the final stage of the module, we employed a weighted summing strategy to integrate the information from different positions in the sequence. Specifically, we first computed the L2 norm for each position in the sequence and converted it to a normalized weight by means of a softmax function. Then, we multiplied the feature vector of each position with its corresponding weight and summed the results to obtain a weighted summation representation. This strategy enabled the model to focus on the more important parts of the sequence and improved the accuracy of the prediction.

### 3.3.3. The pairwise interaction prediction module

The interaction prediction, after processing through the attention mechanism, is followed by four linear layers. They further refine the features extracted from the outputs $o_1$ and $o_2$ (Figure 1E). To prevent overfitting and enhance the model's generalizability, the dropout technique is applied after each linear layer. This regularization method randomly sets a fraction of the input units to zero during the training phase, which makes the model robust to unseen data. The formula for each linear layer can be expressed as follows:

$$p' = \text{LayerNorm}(\text{Dropout}(WH + b)) \quad (11)$$

where W and b are the learnable weight matrices and bias, respectively.

Finally, we jointly optimized all learnable parameters by backpropagation and the Adam optimizer to minimize the binary cross-entropy loss:

$$\mathcal{L} = -\sum_i (y_i \log(p_i) + (1 - y_i)\log(1 - p_i)) \quad (12)$$

where $y_i$ is the ground-truth label of the *i*-th toxin-target pair, and $p_i$ is its output probability from the model.

### 3.4. Evaluation metrics

To evaluate the quality of the classification model for the toxin-target interaction prediction task, several commonly used statistical measurements are employed in this work, including sensitivity (Sn), specificity (Sp), precision (Pre), accuracy (Acc), Matthew's correlation coefficient (MCC) and F1 scores. They are defined as follows:

$$Sn = \frac{TP}{TP + FN}$$

$$Sp = \frac{TN}{TN + FP}$$

$$Pre = \frac{TP}{TP + FP}$$

$$Acc = \frac{TP + TN}{TP + TN + FP + FN} \quad (13)$$

$$F1\ score = 2 \times \frac{Pre \times Sn}{Pre + Sn}$$

$$MCC = \frac{TP \times TN - FP \times FN}{\sqrt{(TP + FP) \times (TP + FN) \times (TN + FP) \times (TN \times FN)}}$$

where TP denotes the count of TTI accurately identified, while TN refers to the correct classification of non-interacting samples. Conversely, FN is the tally of interacting pairs misclassified, and FP is the count of non-interacting samples erroneously identified. Sn quantifies the proportion of true positives accurately classified, whereas Sp similarly measures the true negatives. Pre reflects the fraction of true positives among the predicted positives. F1 score is a harmonic mean that encapsulates both precision and recall, providing a singular measure of test accuracy. MCC offers a balanced measure of the quality of binary classifications.

Additionally, the Receiver Operating Characteristic (ROC) curve, along with the Area Under the Curve (AUC), gauges overall model performance. A ROC curve that skews towards the upper left indicates superior performance, with an AUC approaching 1, signifying near-perfect prediction.

### 3.5. Implementation of tipFormer

The tipFormer was implemented in PyTorch. Specifically, we set the hidden dimension to 32. The model was trained using the LookAhead optimizer [63] combined with the Rectified Adam (RAdam) optimization algorithm with a learning rate of 1e-4. The number of attention heads was set to eight, and the batch size to 1. To mitigate the risk of overfitting, we set the dropout rate to 0.2. Supplementation Table S6 describes all tipFormer's hyperparameters.

### 3.6. Benchmark machine learning models compared

Three conventional machine learning-based and one deep learning-based methods were used to compare with tipFormer on the same dataset. Each model underwent rigorous hyperparameter optimization to ensure optimal performance, as described in Supplementation Table S6. Specifically, we employed random forests (RF), support vector machines (SVM) and *k*-nearest neighbors (KNN) algorithms. When using RF for feature extraction, we set the number of trees to 50 across all computational tests. We set SVM's regularization parameter (C) to 1e-3 and the kernel coefficient (gamma) to 1e-2 and utilize a polynomial (poly) kernel. For KNN, we uniformly set the number of neighbors to 5 throughout the computational tests.

We adopted DeepCNN, a model that employs only CNNs, to extract features from toxins and targets independently. These features are then concatenated and fed to linear layers, without attention mechanisms, to learn interaction information for prediction. Detailed hyperparameters of the model can be found in Supplementation Table S6.

### 4. Conclusion

We developed tipFormer, a transformer-based deep learning approach designed to predict interactions between atmospheric pollutant toxins and human proteins. Our work was built upon the thesis that toxic aerosol pollutants exert their adverse effects on human health through interaction with proteins to alter their functions on key biological processes and signaling pathways. The final tipFormer model can be used as a filter to identify toxic components that can cause complex diseases. By leveraging pre-trained language models for proteins and chemicals and introducing a new cross-attention mechanism, tipFormer can effectively capture the complex interactions between toxins and their targets.

The comprehensive evaluation demonstrates tipFormer's superior performance compared to conventional machine learning and deep learning methods. In addition, tipFormer exhibits generalizability in handling previously unknown or overlooked toxins, targets, or both. This capability is crucial for modernizing and digitally transforming ecological and human health risk assessments across a broad range of environmental applications, including the evaluation of chemicals in air, water, food, and consumer products.

Our model can screen for potential key toxic components in pollutants by evaluating the interactions between compounds in air pollutants and human proteins. This provides high-throughput information on suspect compounds for effect-based pollutant screening. It offers a scientific basis for pollutant control and the discovery of new pollutants. By accurately predicting TTIs, it can help identify aerosol compounds in PM pollutants that may disrupt cellular processes, leading to health problems. The gene-disease association analysis revealed that neurodegenerative diseases of the central nervous system showed highly significant *p*-values, indicating a potential link between toxins from biomass combustion and neurological disease development. These findings are concordant with previous human and animal studies.

Moving forward, we aim to unravel the deeper narratives of toxin impacts on public health and refine tipFormer further. These endeavors promise to propel computational toxicology forward and pave the way for innovative therapeutic interventions aimed at disease prevention and treatment. In addition, we emphasize that tipFormer is a scalable framework that can easily incorporate more information into the current prediction pipeline.

*Competing interests*

The authors declare no competing interests.

*Funding*

The work was supported in part by funding from the Hong Kong RGC Theme-based Research Scheme (TRS grant T24-508/22-N), the Hong Kong RGC Strategic Target Grant (RGC grant STG1/M-501/23-N), the Hong Kong General Research Fund (GRF 15213922), the Hong Kong Global STEM Scholar Scheme, the Hong Kong Jockey Club Charity Trust, and a Hong Kong Polytechnic University Graduate Fellowship.

*Author contributions*

WZ conceived and supervised the project, and YZ and WZ designed the overall experiments and wrote the manuscript. YZ collected data and designed, implemented and tested the tipFormer method and software. YZ and SQ performed function analysis. YH, YL, JL and XL provided experimental data and helped interpret the results.

**Figure 1**. **The overall architecture of tipFormer for TTI Prediction**. (**A**) Data collection: Retrieve data from the T3DB database, serving as the foundation for the tipFormer model. (**B**) Preprocessing: Select high-quality organic toxin molecule samples. (**C**) The encoding module: Integrate pre-trained large language models of toxins and proteins with convolutional neural networks to encode and extract intricate features from proteins and toxins. (**D**) The interaction learning module: Employ a cross-domain attention mechanism to analyze and learn the complex interplay between protein and toxin features. (**E**) The pairwise interaction prediction module: Apply the learned features to predict potential toxin-protein interactions, providing insights into the biological effects of toxins on cellular functions. (**F**) Model evaluation and interpretation: Assess the interpretability and generalizability of the model through performance evaluation and interpretability of predictions.

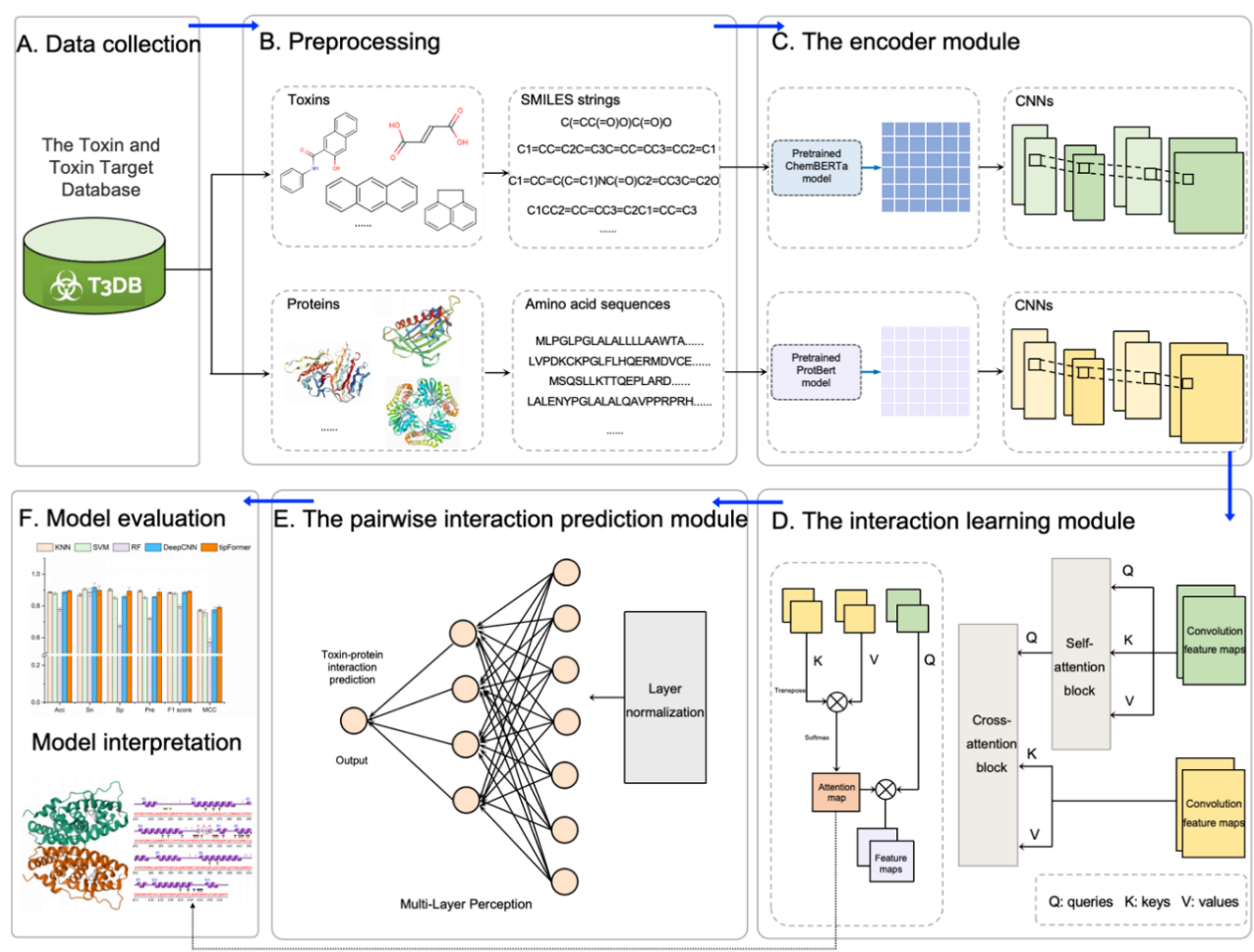

**Figure 2. Evaluating tipFormer using benchmark dataset. (A)** Performance metrics of the five models on the test set. **(B)** ROC curves and average AUC value for five times of tipFormer. **(C)** Raw data feature distribution and **(D)** Final layer feature visualization via t-SNE scatter plots. **(E)** Performance metrics achieved by five models under two new toxin settings. **(F)** Performance metrics achieved by five models under the new-target setting.

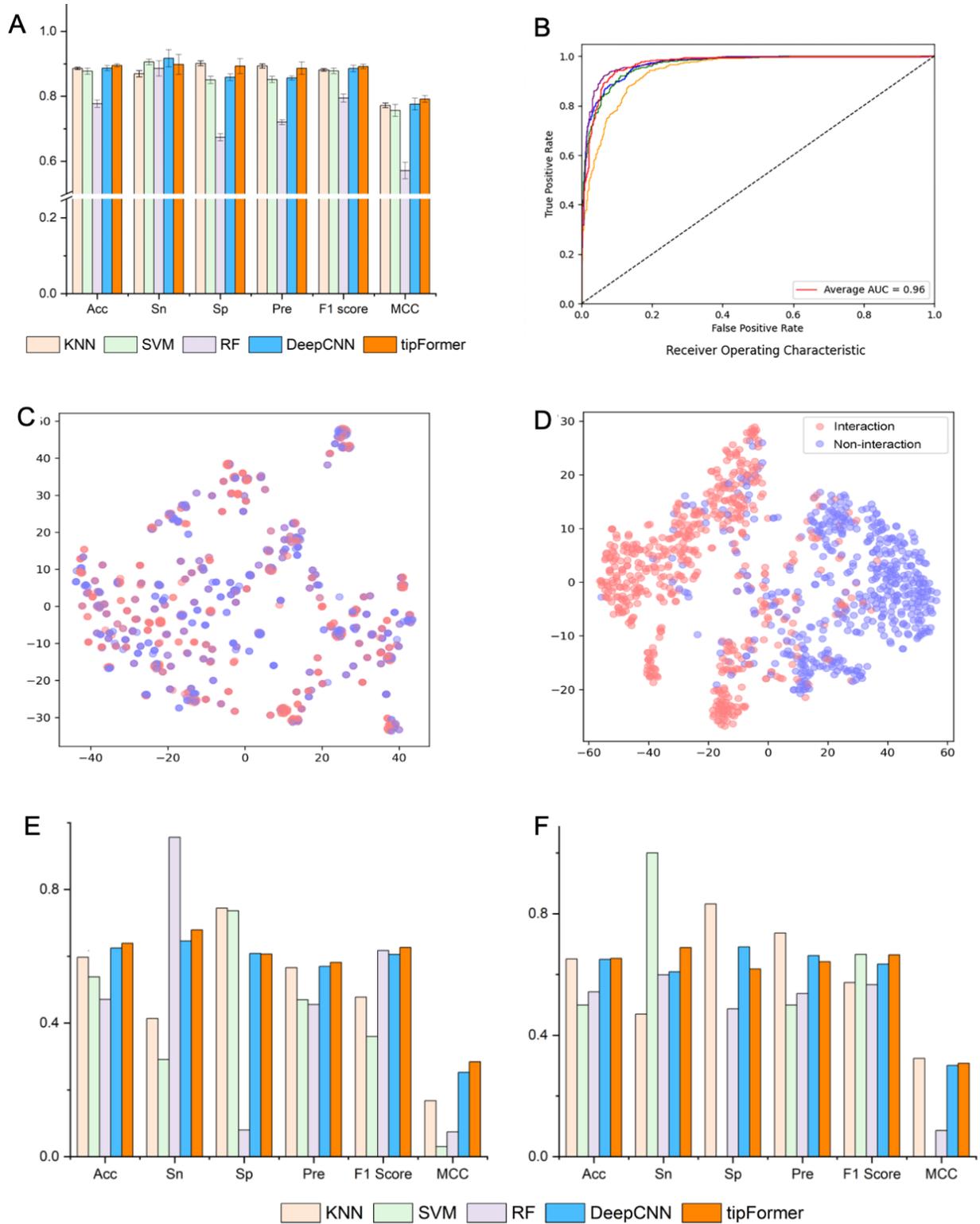

**Figure 3. Molecular configuration and interaction of Bisphenol A with estrogen-related receptor gamma.**
**(A)** The 2D structure graph of the molecular configuration of Bisphenol A. **(B)** A graph illustration of the interaction of Bisphenol A and estrogen-related receptor gamma, highlighting their binding interface. **(C)** A graph of protein amino acid residues shows that the amino acid residues marked with red dots on top interact with Bisphenol A.

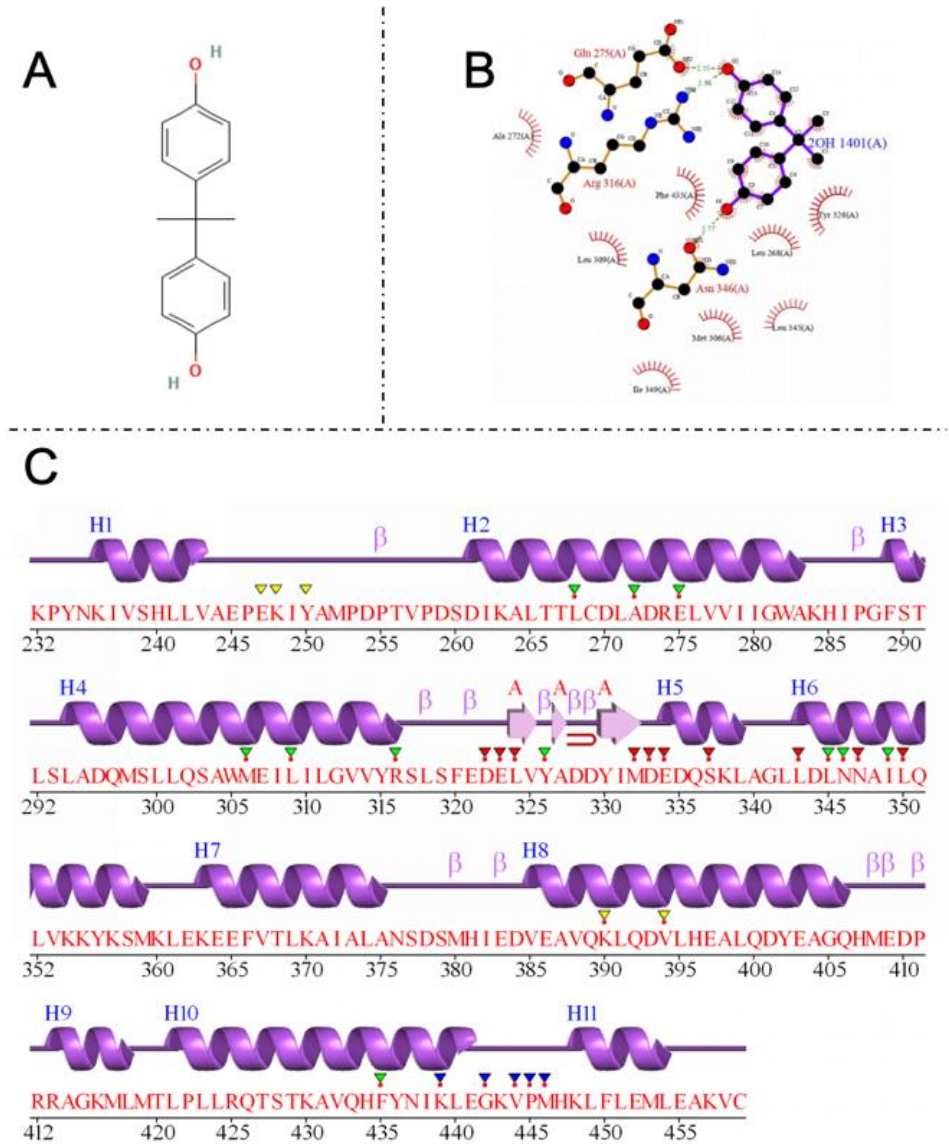

**Table 1.** Comparison of the experimental results of five different methods on the test set. The best result for each measure is highlighted in bold, and the second best is underlined.

| Method | Acc | Sn | Sp | Pre | F1 Score | MCC |
|---|---|---|---|---|---|---|
| KNN | 0.886 (0.004) | 0.870 (0.010) | **0.902 (0.008)** | **0.893 (0.006)** | 0.881 (0.004) | 0.772 (0.007) |
| SVM | 0.878 (0.009) | <u>0.906 (0.009)</u> | 0.850 (0.011) | 0.852 (0.010) | 0.878 (0.009) | 0.757 (0.018) |
| RF | 0.777 (0.011) | 0.886 (0.024) | 0.674 (0.011) | 0.720 (0.008) | 0.795 (0.012) | 0.571 (0.026) |
| DeepCNN | <u>0.887 (0.008)</u> | **0.918 (0.027)** | 0.859 (0.011) | 0.857 (0.006) | <u>0.886 (0.010)</u> | <u>0.776 (0.018)</u> |
| tipFormer | **0.895 (0.005)** | 0.898 (0.031) | <u>0.893 (0.023)</u> | <u>0.886 (0.019)</u> | **0.892 (0.007)** | **0.792 (0.011)** |

**Table 2.** Performance evaluation for five models on the external test set. The best result for each measure is highlighted in bold, and the second best is underlined.

| Method | Pre | TP | FP |
|---|---|---|---|
| KNN | 0.203 | 28 | 110 |
| SVM | **1.000** | **138** | **138** |
| RF | 0.572 | 79 | 59 |
| DeepCNN | 0.754 | 104 | 34 |
| tipFormer | <u>0.790</u> | <u>109</u> | <u>29</u> |

**Table 3.** Gene-disease association analysis of cell membrane proteins using GeDiPNet with statistical significance of *p*-value no greater than 0.05.

| Index | Name | *P*-value |
|---|---|---|
| 1 | Nephrogenic Diabetes Insipidus | 0.021 |
| 2 | Cortical Cataract | 0.021 |
| 3 | Nocturnal Enuresis | 0.021 |
| 4 | Nonepidermolytic Palmoplantar Keratoderma | 0.021 |
| 5 | Toxic Shock Syndrome | 0.025 |
| 6 | Arsenic Encephalopathy | 0.029 |
| 7 | Choledochal Cyst | 0.038 |
| 8 | Malignant Neoplasm Of Tongue | 0.042 |
| 9 | Laryngostenosis | 0.046 |
| 10 | Myoclonic Astatic Epilepsy | 0.046 |

**Table 4.** Putative diseases due to exposure to PM compounds with statistical significance of *p*-value no greater than 0.05.

| Index | Name | *P*-value |
|---|---|---|
| 1 | Degenerative Diseases, Central Nervous System | 0.010 |
| 2 | Neurodegenerative Disorders | 0.010 |
| 3 | Cryptogenic Tonic-Clonic Epilepsy | 0.010 |
| 4 | Pulmonary Hypertension | 0.011 |
| 5 | Tonic-Clonic Epilepsy | 0.017 |
| 6 | Cortical Cataract | 0.021 |
| 7 | Schizophreniform Disorders | 0.027 |
| 8 | Secundum Atrial Septal Defect | 0.032 |
| 9 | Intrahepatic Cholestasis | 0.038 |
| 10 | Isolated Split Hand-Split Foot Malformation | 0.041 |
| 11 | Hypoplastic Left Heart Syndrome | 0.044 |